\documentclass{article}
\usepackage{ijcai23}

\usepackage{times}
\usepackage{soul}
\usepackage[hidelinks]{hyperref}
\usepackage[utf8]{inputenc}
\usepackage[small]{caption}
\usepackage{graphicx}
\usepackage{amsmath}
\usepackage{amsthm}
\usepackage{amssymb}
\usepackage{booktabs}
\usepackage{algorithm}
\usepackage{algorithmic}
\usepackage{enumitem}
\usepackage{multicol}
\usepackage{multirow}
\usepackage[dvipsnames]{xcolor}

\title{HDFormer: High-order Directed Transformer for 3D Human Pose Estimation}

\author{
Hanyuan Chen$^{1}$\thanks{Denotes equal contributions}\and
Jun-Yan He$^{1*}$\and
Wangmeng Xiang$^{1*}$\thanks{W. Xiang is the corresponding author}\and
Zhi-Qi Cheng$^{2*}$\and \\
Wei Liu$^1$\and
Hanbing Liu$^3$\and
Bin Luo$^1$\and
Yifeng Geng$^1$\and
Xuansong Xie$^1$
\affiliations
$^1$Alibaba Group, \quad
$^2$Carnegie Mellon University, \quad
$^3$Tsinghua University
\emails
\{hanyuan.chy, leyuan.hjy, wangmeng.xwm, luwu.lb, cangyu.gyf\}@alibaba-inc.com \\
 zhiqic@cs.cmu.edu, ustclwwx@gmail.com, liuhb21@mails.tsinghua.edu.cn, xingtong.xxs@taobao.com 
 }

\begin{document}
\maketitle

\begin{abstract}
Human pose estimation is a challenging task due to its structured data sequence nature. Existing methods primarily focus on pair-wise interaction of body joints, which is insufficient for scenarios involving overlapping joints and rapidly changing poses. To overcome these issues, we introduce a novel approach, the \textit{\textbf{H}igh-order} \textit{\textbf{D}irected} \textit{Transformer} (HDFormer), which leverages high-order bone and joint relationships for improved pose estimation. Specifically, HDFormer incorporates both self-attention and high-order attention to formulate a multi-order attention module. This module facilitates first-order "joint$\leftrightarrow$joint", second-order "bone$\leftrightarrow$joint", and high-order "hyperbone$\leftrightarrow$joint" interactions, effectively addressing issues in complex and occlusion-heavy situations.~In addition, modern CNN techniques are integrated into the transformer-based architecture, balancing the trade-off between performance and efficiency. HDFormer significantly outperforms state-of-the-art (SOTA) models on Human3.6M and MPI-INF-3DHP datasets, requiring only 1/10 of the parameters and significantly lower computational costs. Moreover, HDFormer demonstrates broad real-world applicability, enabling real-time, accurate 3D pose estimation.\footnote{The source code is in https://github.com/hyer/HDFormer}
\end{abstract}


\section{Introduction}
Despite significant strides in deep learning-based 3D pose estimation~\cite{Iskakov_2019_ICCV,Qiu_2019_ICCV,pavllo2019-3d,li2020-cascaded,zhu2021-posegtac,2021PoseAug,ye2022-faster}, achieving stable, accurate pose sequences remains elusive. The prevalent 3D pose estimation framework takes 2D pose detection results~\cite{Chen2018CPN,SunXLW19} as inputs and estimates depth information via end-to-end Graph Convolutional Networks (GCNs)\cite{cai2019-exploiting,pavllo2019-3d} or Transformers\cite{ZhangCVPR22MixSTE}. However, complex scenarios involving overlapping keypoints, rapid pose changes, and varying scales pose challenges to the depth estimation of 3D keypoints.

\begin{figure}[!t]
    \centering
    \centerline{\includegraphics[width=0.9\linewidth]{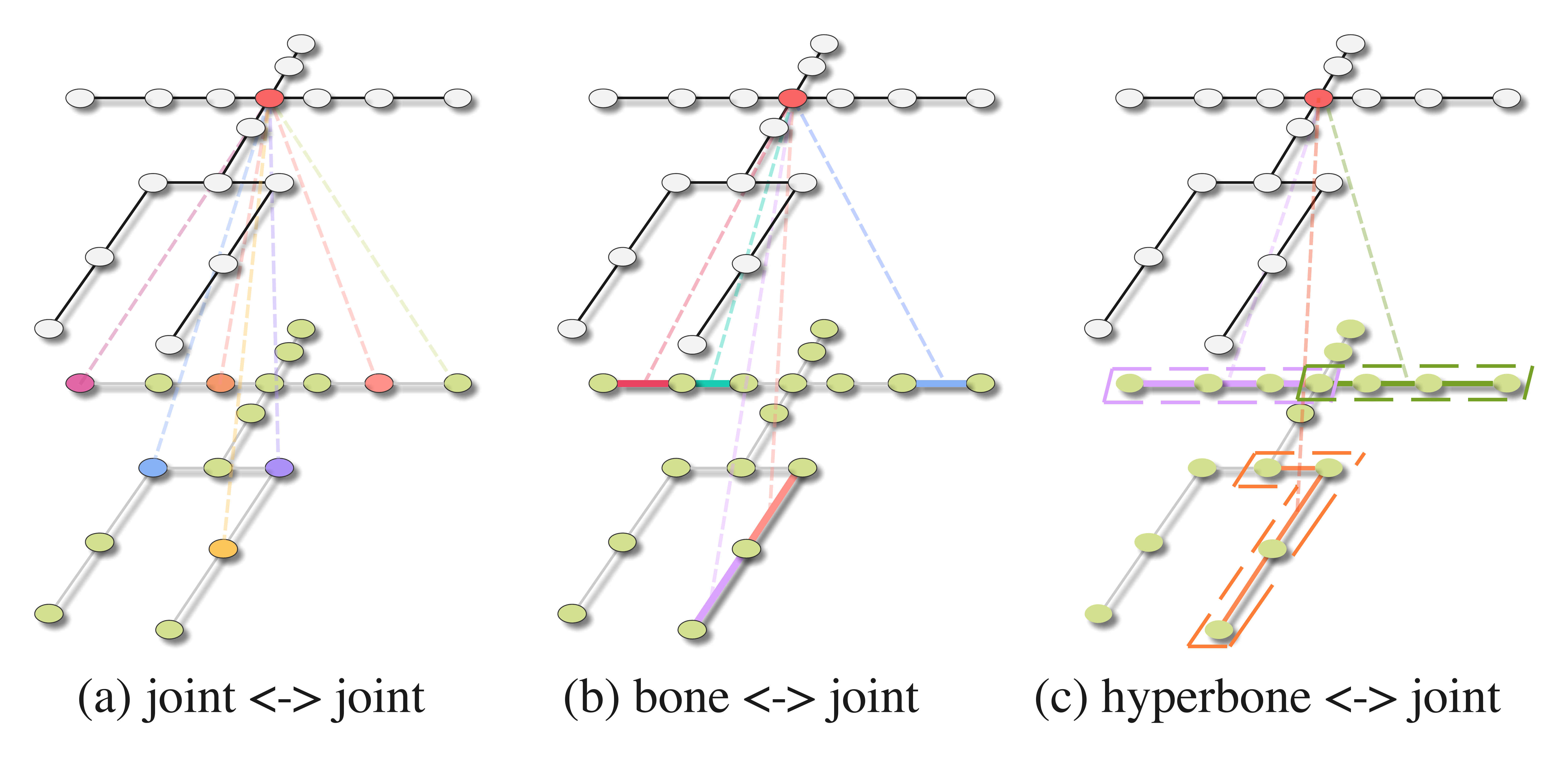}}
    \vspace{-2mm}
    \caption{\small Illustration of first-order (joint$\leftrightarrow{}$joint) attention, second-order (bone$\leftrightarrow{}$joint) and high-order (hyperbone$\leftrightarrow{}$joint) attention. The first-order attention models the connections between joints, while the second-order attention focuses on the relationship between joints and bones. The high-order attention, on the other hand, further delves into the intricate relationships between joints and hyperbones.}
    \vspace{-4mm}
    \label{fig:fron_image}
\end{figure}

Face with these challenges, existing methods mainly utilize first-order "joint $\leftrightarrow$ joint" and second-order "bone $\leftrightarrow$ joint" connections, often overlooking high-order interactions among joint sets (referred to as hyperbones).
Different from first-order (``joint $\leftrightarrow$ joint") and second-order (``bone $\leftrightarrow$ joint") that focus on pair-wise joints/bones connections, high-order relations could describe complex motion dynamics. The high-order interactions contain rich semantic information in motions, as skeletons often move in specific patterns and involve multiple joints and bones simultaneously. 

Learning high-order information without expensive costs is a challenging problem in 3D pose estimation. To address this issue, we propose a novel framework named \textbf{H}igh-order \textbf{D}irected Transformer (HDFormer), which coherently exploits the multi-order information aggregation of skeleton structure for 3D pose estimation.
Specifically, HDFormer leverages the first-order attention to learn the spatial semantics among ``joint $\leftrightarrow$ joint" relationships.
Additionally, it integrates a robust high-order attention module to enhance 3D pose estimation accuracy by capturing both second-order and high-order information.
To encode the hyperbone features, the hyperbone representation encoding module is employed under the constraints of a pre-defined directed human skeleton graph. 
With innovative designs and modern deep learning techniques, HDFormer strikes a commendable balance between efficacy and efficiency. 
In summary, the key contributions of this paper are summarized as follows:

\begin{itemize}[leftmargin=*]
\item We investigate high-order attention module to learn both the \textit{``bone$\leftrightarrow$joint"} and \textit{``hyperbone$\leftrightarrow$joint"} with an effective and efficient cross-attention mechanism. To the best of our knowledge, it is the first end-to-end model to utilize high-order information on a directed skeleton graph for 3D pose estimation.

\item We propose a novel High-order Directed Transformer (HDFormer) for 3D pose estimation. It utilizes \textit{``joint$\leftrightarrow$joint"}, \textit{``bone$\leftrightarrow$joint"} and \textit{``hyperbone$\leftrightarrow$joint"} information with a three-stage U-shape architecture design, which endows the network with the ability to handle more complex scenarios.

\item HDFormer is evaluated on popular 3D pose estimation benchmarks Human3.6M and MPI-INF-3DHP with analysis of quantitative and qualitative results. Specifically, it achieves 21.6\% (96 frames) on Human3.6M without using any extra data, which outperforms the existing SOTA work MixSTE \cite{ZhangCVPR22MixSTE} with only 1/10 parameters and a fraction of computational cost.
\end{itemize}

\section{Related Work}
\subsection{3D Human Pose Estimation}
Despite significant progress in 2D human pose estimation, its 3D counterpart remains a challenge, primarily due to depth ambiguity. Some methods rely on multi-view images/videos \cite{Iskakov_2019_ICCV,Qiu_2019_ICCV,ye2022-faster,he2021db,zhao2018multi,huang2021generating}. However, these setups can be complex and costly, rendering 3D pose estimation with monocular images/videos more practical. Recent studies \cite{pavllo2019-3d,zhu2021-posegtac} have demonstrated the efficiency of lifting 2D joint locations to 3D positions, rather than directly inferring 3D human poses from monocular images/videos. 
To enhance the generalization ability to new datasets or unseen cases, strategies such as data augmentation have been explored to generate diverse, realistic 2D-3D pose pairs, boosting the generalization capability \cite{2021PoseAug,li2020-cascaded}.

\subsection{Graph ConvNet Based Methods}
GCNs~\cite{scarselli2008graph,gilmer2017neural}, extending conventional convolution operators to graphs, have been utilized in several human-related tasks like action recognition \cite{shi2019skeleton,shi2019-twostream,xiang2022language}, action synthesis \cite{yan2019convolutional}, and 3D human pose estimation~\cite{zhou2022hypergraph}. \cite{wang2020-motion} created a lightweight, efficient U-shaped model to capture temporal dependencies. Inspired by \cite{shi2019skeleton}, \cite{hu(2021)-conditional} proposed a conditional directed graph convolution for adaptive graph topology, enhancing non-local dependence. Other methods, such as Semantic Graph Convolution (SemGConv) \cite{zhao2019semantic} and Modulated-GCN \cite{zou2021-modulated}, also prioritize spatial joint relationships. However, they often overlook edge information, particularly high-order relationships. Unlike Skeletal-GNN \cite{zeng2021learning} which utilizes GCN to capture action-specific poses, we establish joint and hyperbone relationships via cross-attention in the directed graph skeleton.

\subsection{Transformer Based Methods}
Transformers, first introduced by \cite{vaswani2017attention}, have been widely used in visual tasks \cite{zhou2022hypergraph,tu2023implicit,cheng2022gsrformer}. In the domain of 3D pose estimation, \cite{li2022-exploiting} proposed the Stride Transformer to lift a long 2D pose sequence to a single 3D pose. PoseFormer \cite{zheng(2021)-poseformer} created a model comprising Spatial-Temporal Transformer blocks to capture spatial and global dependencies across frames. Similarly, MixSTE \cite{ZhangCVPR22MixSTE} captured the temporal motion of body joints over long sequences. While CrossFormer \cite{hassanin2022crossformer} and GraFormer \cite{zhao2022-graformer} encoded dependencies between body joints, our method applies a cross-attention mechanism to integrate high-order relationships and explore the hyperbone-joint relationship. Although transformer-based methods effectively capture global spatial-temporal joint relationships, their computational costs significantly exceed those of GCN-based methods.

\begin{figure*}[!t]
    \centering
\centerline{\includegraphics[width=\textwidth]{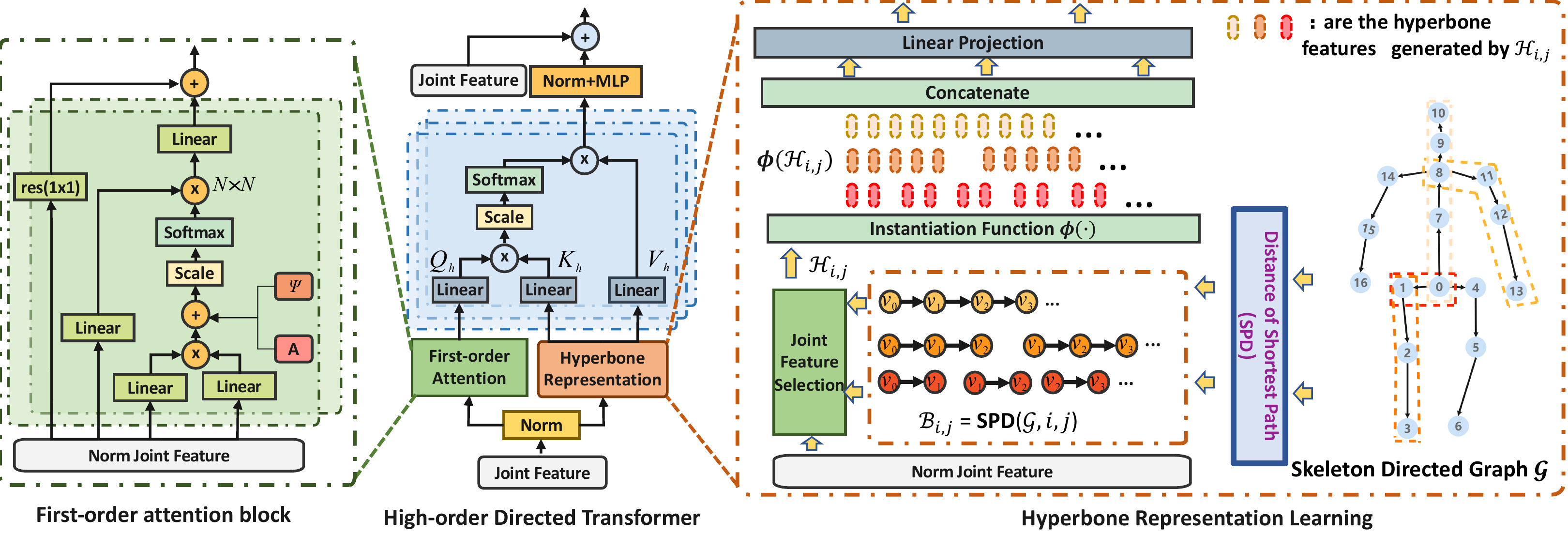}}
    \caption{\small The illustration of High-order Direction Transformer (HDFormer) block. HDFormer block consists of three major parts: (a) First-order attention block to capture ``\textit{joint$\leftrightarrow$ joint}" spatial relationship; (b) Hyperbone representation learning module to encode hyperbone features; (c) High-order attention block to capture both second-order ``\textit{bone $\leftrightarrow$ joint}" and high-order ``\textit{hyperbone$\leftrightarrow$ joint}" interactions. }
    \label{fig:higher order attention block}
\end{figure*}

\section{Proposed Method}
\subsection{Preliminaries}
\noindent \textbf{Directed Skeleton Graph.} The directed skeleton graph $\mathcal{G}$ shown in the right part of Figure \ref{fig:higher order attention block} represents the human skeleton structure, where the nodes are human skeleton joints and the arrows are human skeleton bones. Generally, the human skeleton can be represented as a graph $\mathcal{G} = (\mathcal{V}; \mathcal{E})$, where the vertices are human skeleton joints and edges are physical connections between two joints. Here $\mathcal{V}$ is the set of $N$ joints and $\mathcal{E}$ is characterized by the adjacency matrix $ \bf{A} \in \mathbb{R}^{\it{N \times N}}$. 
The raw pose data, i.e., the joint keypoints vector, is a set of 2D coordinates. In this way, the pose data is transformed into a graph sequence and specifically represented as a tensor $ \bf{X} \in \mathbb{R}^{\it{T \times N \times C}}$, where $T$, $N$, and $C$ denote the temporal length, numbers of joints and channels, respectively. We use the directed graph because it allows for a convenient hyperbone definition.

\noindent \textbf{Transformer.}~Our model's attention mechanism is built upon the original implementation of the classic Transformer \cite{vaswani2017attention}. The attention computing with \textit{query}, \textit{key} and \textit{value} matrix $Q$, $K$, $V$ in each head are depicted as:
\begin{equation}
    \text{Attn}(Q,K,V) = \text{Softmax}((QK^T + \textbf{A} + \Psi) /\sqrt{d_m})V,
    \label{eq:attn}
\end{equation}
where ${Q,K,V} \in \mathbb{R}^{\it{N \times d_m}}$, $N$ is the number of tokens, and $d_m$ indicates the dimension of each token. The multi-head attention of $S$ heads is defined as follows:
\begin{equation}
    \hbar_i = \text{Attn}(Q_i,K_i,V_i), \textit{i} \in \{\textit{1},\dots,\textit{S}\},
\end{equation}
\begin{eqnarray}
    \text{MSA} = \text{Concat}(\hbar_1, ..., \hbar_S)W_o,
    \label{eq:MHSA_concat}
\end{eqnarray}
where $\hbar$ is the attention calculation result for a single head, $ W_o \in \mathbb{R}^{\it{d_m \times d_m}}$ is the linear projection weight. $\bf{A}$ is the adjacency matrix and $\Psi$ is a learnable adjacency matrix. The matrix $\bf{A}$ is fixed and represents the predetermined connections between joints, while the learnable adjacency matrix $\Psi$ adjusts the connection weights based on the input data, improving the capturing of spatial relationships between different joints. The ablation study on the impact of $\Psi$ is presented in line 1 of Table \ref{table:H36M_GT2d_ablation}.

\subsection{High-order Directed Transformer}
\label{sec:High-order Directed Transformer}
The spatial connections between ``joint$\leftrightarrow$joint" and ``joint$\leftrightarrow$bone" are referred to as first-order and second-order information in the 3D pose estimation, which is widely studied in the previous works \cite{ZhangCVPR22MixSTE}. Nevertheless, pairwise first-order and second-order information alone cannot fully describe the complex human skeleton dynamics in the 2D to 3D mappings. For example, human skeletons often move in specific patterns and involve multiple joints and bones at the same time. This observation leads us to further investigate the high-order information interaction of the human skeleton by integrating the high-order attention learning with directed graph and propose a High-order Direct Transformer (HDFormer) for 3D pose estimation.

\noindent \textbf{First-order Attention Modeling.} 
The joint sets of the skeleton describe the rough posture of the human body, and the global multi-head attention~\cite{ZhangCVPR22MixSTE} has demonstrated its effectiveness in 3D pose estimation. Therefore, the multi-head attention scheme is adopted in this work for first-order attention modeling. As illustrated in Figure~\ref{fig:higher order attention block}(a), $ A \in \mathbb{R}^{\it{N \times N}}$ is the adjacency matrix and  $\Psi$ is the learnable adjacency matrix which has the same dimension as $A$. Specifically, given the joint token set $\mathcal{Z} = \{z_{1}\dots\ z_{i}\}$ where $i \in N$ denotes the index of skeleton joints, $z_{i} \in \mathbb{R}^{C}$, $C$ represents the feature channel. The first-order self-attention modeling and feature of joints can be obtained by following Eq.~\ref{eq:attn}, where \textit{query}, \textit{key}, and \textit{value} matrices are generated by three linear layers, $Q=W_{q}\mathcal{Z}$, $K=W_{k}\mathcal{Z}$, and $V=W_{v}\mathcal{Z}$, respectively. $S$ represents the number of heads and $\hbar_{k}$ is the output of each head. Unlike the traditional multi-head attention fuses the output of the attention module with concatenating (Eq.~\ref{eq:MHSA_concat}), we revamp the fusion scheme with a simple accumulation:
\begin{equation}
    \hat{\mathcal{Z}} = \sum_{k=1}^{S} \hbar_{k},
\end{equation}
where $\hat{\mathcal{Z}}$ denote the final output of first-order attention.

\noindent \textbf{Hyperbone Representation.} In this section, we outline the process of constructing and learning the hyperbone representation. A hyperbone is a series of joints and bones that are connected sequentially. The human skeleton can be represented as a special type of graph without loops, allowing for the unique determination of the shortest path between two joints. Given a starting and ending joint, the corresponding hyperbone can be identified using the \textit{distance of the shortest path} (SPD). Specifically, as shown in Figure \ref{fig:higher order attention block}, the human skeleton can be described as a directed graph $\mathcal{G}$. The ``hip" joint is defined as the directed graph's root node. Given two joint nodes on the directed graph, we could follow the direction of edges to find the shortest path from starting joint $i$ to $j$. For example, there is a shortest path from joint index 0 to joint index 3 by \text{[0, 1, 2, 3]}, which is done by moving from index 0 to index 3 following the edges of the directed graph (bone for human skeleton).

Formally, given the human skeleton-directed graph $\mathcal{G}$ and the $(start,end)$ joint indices ($i$, $j$), we could utilize the shortest path algorithm (SPD) to discover the joint set belonging to the hyperbone $\mathcal{B}_{i,j}=\{v_{h_i},v_{h_{i+1}},\dots,v_{h_j} \}$, where $v_{*}$ represents the human joint, $|\mathcal{B}_{i,j}| = n$ represents the number of joints in hyperbone, and we call this the \textit{order} of hyperbone, $h_{*}$ is the joint index:
\begin{equation}
    \mathcal{B}_{i,j} = \text{SPD}(\mathcal{G}, i, j),
\end{equation}

To encode hyperbone features, we propose a novel hyperbone encoding method. Specifically, the feature of hyperbone can be obtained by a function $\phi(\cdot)$ that takes hyperbone joint set $\mathcal{B}_{i,j}$'s corresponding features $\mathcal{H}_{i,j} = \{ z_{h_i}, z_{h_{i+1}},\dots,z_{h_j}\}$ as input and generate hyperbone features, where $z_{h_i}$ is the feature of joint $v_{h_i}$.

\noindent \textbf{Instantiation.} Previous works, e.g. Anatomy \cite{chen(2021)-anatomyaware}, have used a simple subtraction of joint features to construct bone features. In contrast, we propose a general process for constructing both bone and hyperbone features, and offer instantiation methods. Specifically, we investigate several instantiations of the function $\phi(\cdot)$.

\textit{Subtraction.} $\phi(\cdot)$ can be defined as a subtraction operation. As we use a directed graph to represent the human skeleton when adopting subtraction for hyperbone representation, it is equivalent to the subtraction of start and end joints:
\begin{equation}
    \phi ( \mathcal{H}_{i,j} ) = f(z_{h_i} -z_{h_j}),
\end{equation}
where $z$ is the joint feature, $f$ is a linear mapping. This representation is easy to calculate and works fine for second-order bone representation, however, it loses information on bone sequence for hyperbone with higher order.

\textit{Summation/Multiplication.} $\phi(\cdot)$ can also be defined as element-wise summation or multiplication for joints:

\begin{equation}
    \phi ( \mathcal{H}_{i,j} ) = \sum_{z\in \mathcal{H}_{i,j}}{f(z) / n},
\end{equation}
\begin{equation}
    \phi ( \mathcal{H}_{i,j} ) = \prod_{z\in \mathcal{H}_{i,j}}{f(z)},
\end{equation}

\noindent where $n$ is the number of joints, $f$ is a linear mapping.

\textit{Concatenation.} $\phi(\cdot)$ can be defined with concatenation and linear mapping:

\begin{equation}
    \phi ( \mathcal{H}_{i,j} ) = f([z_{h_1},\dots,z_{h_n}]),
\end{equation}

\noindent where the operator $[\cdot]$ represents the concatenation of features in the shortest path, $f$ maps the concatenated feature to the same dimension as the joint feature. 

\textit{Sub-Concat.} To overcome the sequence information loss issue in subtraction, we combine subtraction and concatenation for a mixed function for $\phi(\cdot)$:

\begin{equation}
    \phi ( \mathcal{H}_{i,j} ) = f([z_{h_1} - z_{h_2},\dots,z_{h_{n-1}} - z_{h_n}]),
\end{equation}

\noindent where the second-order bone feature is calculated with subtraction and the high-order hyperbone feature is obtained with concatenation and linear mapping.

\noindent \textbf{High-order Directed Transformer.} Figure \ref{fig:higher order attention block}(b) illustrates the architecture of our proposed High-order Directed Transformer block, which consists of three components: first-order attention block, hyperbone encoding block, and high-order attention block. The cross-attention fusion involves joint features $ \hat{\mathcal{Z}}$ from first-order attention modeling block and hyperbone feature $H=[Y_{2},..Y_{o},..Y_{n}]$, where $Y_o$ represent hyperbone features with order $o$ from hyperbone encoding block. Formally, the cross-attention fusion can be expressed as:

\begin{equation}
\begin{aligned}
        Y_o &= [\phi(\mathcal{H}_{i,j})], |\mathcal{H}_{i,j}| = o, \\
        H &= [Y_2,\dots,Y_n], \\
        Q_h = W_{q_h}  \hat{\mathcal{Z}},
        K_h &= W_{k_h} H,
        V_h = W_{v_h} H, \\
        \text{CrossAttn}(Q_h,K_h,V_h) &= \text{Softmax}(Q_hK_{h}^T/\sqrt{d_m})V_{h}, \\
\end{aligned}
\end{equation}

\noindent where $W_{q_h}, W_{k_h}, W_{v_h}$ are learnable parameters. Since we only use the joint feature in the query, the computation and memory complexity of generating the cross-attention map in cross-attention are linear rather than quadratic as in all-attention, making the entire process more efficient. Similar to MHSA~\cite{vaswani2017attention}, we also adopt a multi-head attention design and add an MLP layer after the attention layer.

\begin{figure*}[!t]
    \centering
    \centerline{\includegraphics[width=\textwidth]{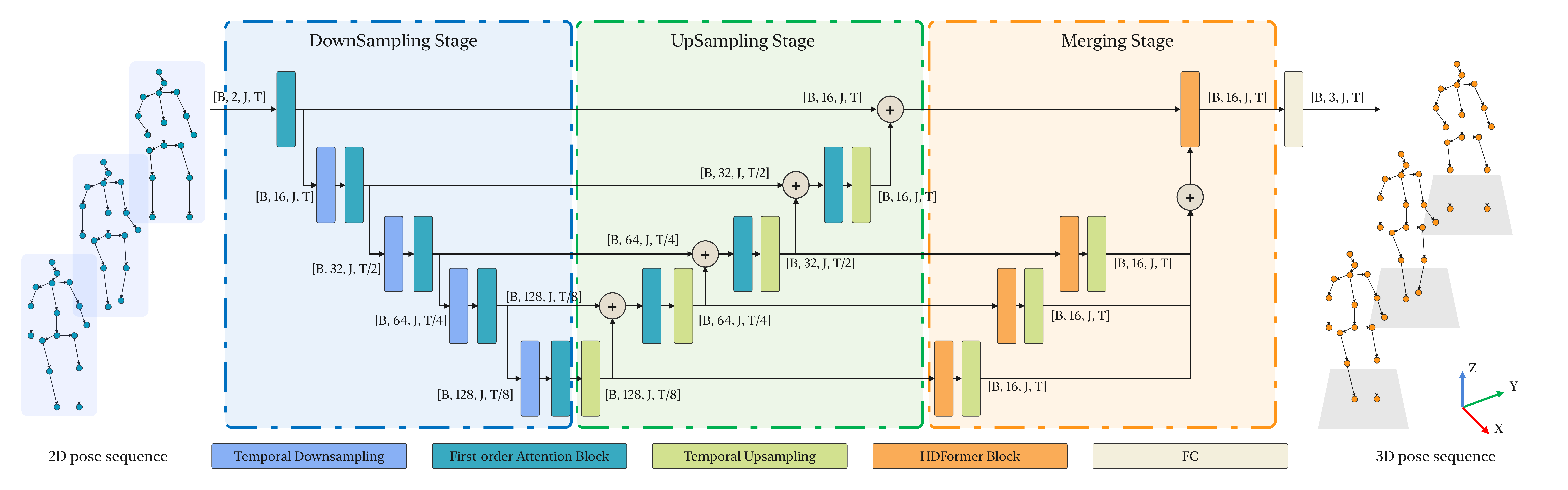}}
    \vspace{-2mm}
    \caption{\small Overview of our framework: A High-order Directed Transformer with a U-shaped design for 3D human pose estimation. The framework includes downsampling, upsampling, and merging stages, incorporating high-order attention and multi-scale temporal information.}
    \vspace{-2mm}
    \label{fig:arch}
\end{figure*}

\noindent \textbf{Loss Function.}~We adopted the loss function similar to UGCN \cite{wang2020-motion}, which was formulated as follow:
\begin{equation}
    \mathcal{L} = \mathcal{L}_p + \lambda \mathcal{L}_m,
\end{equation}

\noindent where $\mathcal{L}_p$ is the 3D joint coordinates loss, which is defined as the mean per joint position error (MPJPE), and $\mathcal{L}_m$ is motion loss introduced by \cite{wang2020-motion}, $\lambda$ is a hyperparameter for balancing two objectives and is set to 0.1. Motion loss allows our model to capture more natural movement patterns of the keypoints in the prediction, since Minkowski Distance loss does not consider the similarity of temporal structure between the estimated pose sequence and ground truth.

\subsection{Network Architecture}
To achieve a satisfactory trade-off between effectiveness and efficiency, we construct the network architecture as a U-shaped transformer network using the aforementioned high-order transformer block to learn the multi-order information. 

As illustrated in Figure \ref{fig:arch}, the proposed network architecture contains three stages: 1) Downsampling stage collects long-time range information by temporal pooling. The temporal downsampling block has an inside temporal convolution’s stride set to 2 and kernel size set to 5. It is used to downsample the temporal resolution; 2) Upsampling stage recovers the temporal resolution, and skip connections are adopted between the downsampling stage and the upsampling stage to integrate the low-level features. The temporal upsampling block is the conventional bilinear interpolation along the temporal axis to recover higher temporal resolution. 3) Merging stage transforms the feature maps at different temporal scales in the upsampling stage, and fuses them to obtain the final embedding. Finally, the 3D coordinate for each keypoint is regressed by a fully connected layer. Overall, our model takes 2D human poses estimated by an off-the-shelf 2D human pose estimator as input, and generates the corresponding 3D poses. Our model is trained with motion loss \cite{wang2020-motion} in an end-to-end manner. The feature dimensions of our network are shown in Figure \ref{fig:arch}, where the batch size, the number of nodes, and the sequence length are represented by symbols B, J, and T, respectively.

As demonstrated in Section \ref{sec:High-order Directed Transformer}, the \textit{First-order Attention} (FOA) blocks can utilize the pair-wise spatial and temporal relationship between joints, whereas the \textit{High-order Attention} (HOA) blocks further construct the complex feature interaction between joints and hyperbones at different scales. To balance the stability and complexity, we adopt FOA blocks in the downsampling and upsampling stages while adopting HOA blocks in the merging stage. This configuration is further discussed with an ablation study as shown in Table \ref{table:ablation_stages}.

\section{Experiment}
\subsection{Datasets and Metric}
Experiments are conducted on the 3D pose estimation benchmark dataset Human3.6M\cite{Ionescu_POSE_TPAMI14} and MPI-INF-3DHP \cite{Mehta_RCFSXT_3DV17}. Human3.6M is the most widely used evaluation benchmark, containing 3.6 million video frames captured from four synchronized cameras with different locations and poses at 50 Hz. 11 subjects are performing 15 kinds of actions. MPI-INF-3DHP is a 3D human body pose estimation dataset consisting of both constrained indoor and complex outdoor scenes.  It consists of 1.3M frames captured from the 14 cameras. For fair comparisons, the evaluation metric MPJPE is adopted in this work, which follows the setting of the previous works~\cite{hu(2021)-conditional,cai2019-exploiting,ZhangCVPR22MixSTE,zhao2022-graformer}. Unlike the 2D pose estimation task, MPJPE is proposed to evaluate models comprehensively for accuracy and stability.

\subsection{Implementation Details}
The proposed HDFormer is implemented with the PyTorch platform and all the experiments are conducted on a single NVIDIA TITAN V100 GPU. We optimized the model by the AdaMod optimizer \cite{ding2019adaptive} for 110 epochs with a batch size of 256, and the base learning rate is $5\times10^{-3}$ with decayed by 0.1 at 80, 90, and 100 epochs. To avoid over-fitting, we set the weight decay factor to $10^{-5}$ for parameters of convolution layers and the dropout rate to 0.3 at part of the layers. Besides, we followed UGCN \cite{wang2020-motion} to apply the sliding window algorithm with a step length of 5 to estimate a variable-length pose sequence with fixed input length at inference time.

\subsection{Quantitative Evaluation}
\begin{table*}
    \scriptsize
    \centering
    \tiny
    \caption{\small Quantitative comparisons with state-of-the-art methods on Human3.6M under protocol \#1 and protocol \#2, where methods marked with $\dag$ are video-based; T denotes the number of input frames; and CPN and HR-Net denote the input 2D poses are estimated by \protect\cite{Chen2018CPN} and \protect\cite{SunXLW19}, respectively. The best results of CPN and HR-Net are marked in \textcolor{red}{red} and \textcolor{blue}{blue}, respectively.}
    \vspace{-0.1in}
    \setlength{\tabcolsep}{2mm}
    {
    \begin{tabular}{l|c|c|c|c|c|c|c|c|c|c|c|c|c|c|c|c}
    \hline
    \textbf{Protocol \#1} & Dir. & Disc & Eat & Greet & Phone & Photo & Pose & Purch. & Sit & SitD. & Smoke & Wait & WalkD. & Walk & WalkT. & Avg. \\ \hline

    Cai \cite{cai2019-exploiting} (CPN, T=7)  & 44.6 & 47.4 & 45.6 & 48.8 & 50.8 & 59.0 & 47.2 & 43.9 & 57.9 & 61.9 & 49.7 & 46.6 & 51.3 & 37.1 & 39.4 & 48.8 \\ 
    Pavllo \cite{pavllo2019-3d} (CPN, T=243)  & 45.2 & 46.7 & 43.3 & 45.6 & 48.1 & 55.1 & 44.6 & 44.3 & 57.3 & 65.8 & 47.1 & 44.0 & 49.0 & 32.8 & 33.9 & 46.8 \\ 
    Xu \cite{xu2020-deep} (CPN, T=9)  & \textcolor{red}{37.4} & 43.5 & 42.7 & 42.7 & 46.6 & 59.7 & 41.3 & 45.1 & 52.7 & 60.2 & 45.8 & 43.1 & 47.7 & 33.7 & 37.1 & 45.6 \\ 
    Liu \cite{liu2020-attention}(CPN, T=243)  & 41.8 & 44.8 & 41.1 & 44.9 & 47.4 & 54.1 & 43.4 & 42.2 & 56.2 & 63.6 & 45.3 & 43.5 & 45.3 & 31.3 & 32.2 & 45.1 \\ 
    Wang \cite{wang2020-motion} (CPN, T=96)   & 41.3 & 43.9 & 44.0 & 42.2 & 48.0 & 57.1 & 42.2 & 43.2 & 57.3 & 61.3 & 47.0 & 43.5 & 47.0 & 32.6 & 31.8 & 45.6 \\ 
    Hu \cite{hu(2021)-conditional} (CPN, T=96)  & 38.0 & 43.3 & 39.1 & \textcolor{red}{39.4} & 45.8 & 53.6 & 41.4 & 41.4 & 55.5 & 61.9 & 44.6 & 41.9 & 44.5 & 31.6 & \textcolor{red}{29.4} & 43.4 \\ 
    Zhang \cite{ZhangCVPR22MixSTE} (CPN, T=81)  & 39.8 & \textcolor{red}{43.0} & \textcolor{red}{38.6} & 40.1 & \textcolor{red}{43.4} & 50.6 & \textcolor{red}{40.6} & 41.4 & \textcolor{red}{52.2} & \textcolor{red}{56.7} & 43.8 & \textcolor{red}{40.8} & 43.9 & \textcolor{red}{29.4} & 30.3 & \textcolor{red}{42.4} \\ 
    Wang \cite{wang2020-motion} (HR-Net, T=96)   & 38.2 & 41.0 & 45.9 & 39.7 & 41.4 & 51.4 & 41.6 & 41.4 & 52.0 & 57.4 & 41.8 & 44.4 & 41.6 & 33.1 & 30.0 & 42.6 \\ 
    Hu \cite{hu(2021)-conditional} (HR-Net, T=96)  & 35.5 & 41.3 & 36.6 & 39.1 & 42.4 & 49.0 & 39.9 & 37.0 & 51.9 & 63.3 & 40.9 & 41.3 & 40.3 & 29.8 & \textcolor{blue}{28.9} & 41.1 \\ 
    Zhang \cite{ZhangCVPR22MixSTE} (HR-Net, T=243)  & 36.7 & \textcolor{blue}{39.0} & 36.5 & 39.4 & \textcolor{blue}{40.2} & \textcolor{blue}{44.9} & 39.8 & \textcolor{blue}{36.9} & \textcolor{blue}{47.9} & \textcolor{blue}{54.8} & \textcolor{blue}{39.6} & \textcolor{blue}{37.8} & \textcolor{blue}{39.3} & 29.7 & 30.6 & \textcolor{blue}{39.8} \\ 
    \hline
    HDFormer(CPN, T=96)  & 38.1 & 43.1 & 39.3 & 39.4 & 44.3 & \textcolor{red}{49.1} & 41.3 & \textcolor{red}{40.8} & 53.1 & 62.1 & \textcolor{red}{43.3} & 41.8 & \textcolor{red}{43.1} & 31.0 & 29.7 & 42.6 \\ 
    HDFormer (HR-Net, T=96)  & \textcolor{blue}{34.7} & 41.7 & \textcolor{blue}{36.0} & \textcolor{blue}{38.4} & 41.1 & 45.3 & \textcolor{blue}{39.6} & 37.4 & 49.0 & 63.1 & 39.8 & 38.9 & 40.2 & \textcolor{blue}{29.3} & 29.1 & 40.3 \\ 

    \hline
    \hline
    \textbf{Protocol \#2}  & Dir. & Disc & Eat & Greet & Phone & Photo & Pose & Purch. & Sit & SitD. & Smoke & Wait & WalkD. & Walk & WalkT. & Avg. \\ 
    \hline
    Cai \cite{cai2019-exploiting} (CPN, T=7)  & 35.7 & 37.8 & 36.9 & 40.7 & 39.6 & 45.2 & 37.4 & 34.5 & 46.9 & 50.1 & 40.5 & 36.1 & 41.0 & 29.6 & 33.2 & 39.0 \\ 
    Pavllo \cite{pavllo2019-3d} (CPN, T=243)  & 34.1 & 36.1 & 34.4 & 37.2 & 36.4 & 42.2 & 34.4 & 33.6 & 45.0 & 52.5 & 37.4 & 33.8 & 37.8 & 25.6 & 27.3 & 36.5 \\ 
    Xu \cite{xu2020-deep} (CPN, T=9)  & 31.0 & 34.8 & 34.7 & 34.4 & 36.2 & 43.9 & 31.6 & 33.5 & 42.3 & 49.0 & 37.1 & 33.0 & 39.1 & 26.9 & 31.9 & 36.2 \\ 
    Liu \cite{liu2020-attention}(CPN, T=243)  & 32.3 & 35.2 & 33.3 & 35.8 & 35.9 & 41.5 & 33.2 & 32.7 & 44.6 & 50.9 & 37.0 & 32.4 & 37.0 & 25.2 & 27.2 & 35.6 \\ 
    Wang \cite{wang2020-motion} (CPN, T=96)   & 32.9 & 35.2 & 35.6 & 34.4 & 36.4 & 42.7 & 31.2 & 32.5 & 45.6 & 50.2 & 37.3 & 32.8 & 36.3 & 26.0 & 23.9 & 35.5 \\ 
    Hu \cite{hu(2021)-conditional} (CPN, T=96)  & 29.8 & 34.4 & 31.9 & 31.5 & 35.1 & 40.0 & \textcolor{red}{30.3} & \textcolor{red}{30.8} & 42.6 & 49.0 & 35.9 & 31.8 & 35.0 & 25.7 & \textcolor{red}{23.6} & 33.8 \\ 
    Zhang \cite{ZhangCVPR22MixSTE} (CPN, T=81)  & 32.0 & 34.2 & \textcolor{red}{31.7} & 33.7 & 34.4 & 39.2 & 32.0 & 31.8 & 42.9 & \textcolor{red}{46.9} & 35.5 & 32.0 & 34.4 & \textcolor{red}{23.6} & 25.2 & 33.9 \\ 
    Wang \cite{wang2020-motion} (HR-Net, T=96)   & 28.4 & 32.5 & 34.4 & 32.3 & 32.5 & 40.9 & 30.4 & 29.3 & 42.6 & \textcolor{blue}{45.2} & 33.0 & 32.0 & 33.2 & 24.2 & 22.9 & 32.7 \\ 
    Hu \cite{hu(2021)-conditional} (HR-Net, T=96)  & \textcolor{blue}{27.7} & 32.7 & 29.4 & 31.3 & 32.5 & 37.2 & 29.3 & 28.5 & 39.2 & 50.9 & 32.9 & 31.4 & 32.1 & 23.6 & 22.8 & 32.1 \\ 
    Zhang \cite{ZhangCVPR22MixSTE} (HR-Net, T=243)  & 28.0 & \textcolor{blue}{30.9} & \textcolor{blue}{28.6} & 30.7 & \textcolor{blue}{30.4} & \textcolor{blue}{34.6} & \textcolor{blue}{28.6} & \textcolor{blue}{28.1} & \textcolor{blue}{37.1} & 47.3 & \textcolor{blue}{30.5} & \textcolor{blue}{29.7} & \textcolor{blue}{30.5} & \textcolor{blue}{21.6} & \textcolor{blue}{20.0} & \textcolor{blue}{30.6} \\ 
    \hline
    HDFormer (CPN, T=96)  & \textcolor{red}{29.6} & \textcolor{red}{33.8} & 31.7 & \textcolor{red}{31.3} & \textcolor{red}{33.7} & \textcolor{red}{37.7} & 30.6 & 31.0 & \textcolor{red}{41.4} & 47.6 & \textcolor{red}{35.0} & \textcolor{red}{30.9} & \textcolor{red}{33.7} & 25.3 & 23.6 & \textcolor{red}{33.1} \\ 
    HDFormer (HR-Net, T=96)  & 27.9 & 32.8 & 29.7 & \textcolor{blue}{30.6} & 32.5 & 35.0 & 28.9 & 29.2 & 38.3 & 50.0 & 32.9 & 30.1 & 31.8 & 23.6 & 22.8 & 31.7 \\ 

    \hline
    \end{tabular}
    }
    \label{table:H36M_cpn_hrnet}
\end{table*}


\begin{table}
    \scriptsize
    \centering
    \caption{\small Results on Human3.6M with ground-truth 2D poses as input. Our method with subtraction feature representation is marked with *. The latency is measured with batch size = 1.}
    \vspace{-0.1in}
    \setlength{\tabcolsep}{1.35mm}{
    \begin{tabular}{l|c|c|c|c}
    \hline
    Methods            & MPJPE[↓]  & Params & Latency &  Frames \\ \hline
    U-CondDGCN \cite{hu(2021)-conditional}    & 22.7 & 3.4 M & 0.6 ms  & 96\\
    Cai \cite{cai2019-exploiting}     & 37.2 &  5.04M     &    11.6ms   & 7\\
    Pavllo \cite{pavllo2019-3d}       & 37.2 &  17.0M     &   -  &  243\\
    Liu \cite{liu2020-attention}      & 34.7 &  11.25M     &  9.9ms   & 243\\
    Wang \cite{wang2020-motion}     & 25.6 &   1.69M    &   -  &  96\\
    MixSTE \cite{ZhangCVPR22MixSTE}  & 25.9 &  33.7M     &   2.6ms     & 81\\
    MixSTE $^{\ddagger}$\cite{ZhangCVPR22MixSTE}   & 21.6 &  33.8M     &   8.0 ms  & 243\\
\hline
    HDFormer*                        & 22.1 & 2.8 M & 0.9 ms & 96\\
    HDFormer & \textbf{21.6} & 3.7 M & 1.3 ms & 96 \\ \hline
    
    \end{tabular}}
    \label{table:H36M_GT2d}
    \vspace{-2mm}
\end{table}
\noindent \textbf{Results on Human3.6M.} The proposed approach is compared with the state-of-the-art methods to evaluate the performance. In this subsection, the reported performance in their original paper is directly copied as their results.
The performance comparison with the state-of-the-art works on Human3.6M \cite{Ionescu_POSE_TPAMI14} is listed in Table \ref{table:H36M_cpn_hrnet}, including graph ConvNet-based and Transformer-based methods. For fair comparisons with other SOTA methods, we consider not only the effectiveness of the model but also the scale of parameters and latency in Table~\ref{table:H36M_GT2d}, which can comprehensively demonstrate the real-world performance of the model. To our knowledge, this is the first comprehensive comparison experiment on the benchmark dataset Human3.6M. 
The current SOTA method MixSTE \cite{ZhangCVPR22MixSTE}, a transformer-based model, achieves 25.9\% and 21.6\% MPJPE with the input of 81 and 243-frame sequences, respectively. Compared to MixSTE, the proposed HDFormer achieves 21.6\% MPJPE with the input of only 96 frames. More importantly, our model has only a 1/10 scale of 3.7 M vs. 33.8 M and six times the speed. The graph ConvNet-based SOTA method U-CondDGCN has a very small scale of parameters, latency, and ideal performance. However, the proposed HDFormer is a transformer-based method that achieves significant improvement compared to U-CondDGCN with a very close scale of parameters and same-level latency.  The comparisons powerfully demonstrate the effectiveness and efficiency of HDFormer. 

\noindent \textbf{Results on MPI-INF-3DHP.} In Table \ref{table:MPI}, we compared our method with state-of-the-art methods on the MPI-INF-3DHP benchmark to evaluate the generalization ability of the proposed HDFormer. We take the ground-truth 2D poses as model input. Our method achieves the same trends as the results on Human3.6M, which is also the SOTA performance under the metric of PCK, AUC, and MPJPE.

\begin{table}
    \scriptsize
    \centering
    \caption{\small Results on MPI-INF-3DHP with three metrics.}
    \vspace{-0.1in}
    \setlength{\tabcolsep}{1.35mm}{
    \begin{tabular}{l|c c c}
    \hline
    Methods        & PCK[↑] & AUC[↑] & MPJPE[↓] \\ \hline
    CNN \cite{mehta2017-monocular}   & 75.7 & 39.3 & - \\
    VNect(ResNet101) \cite{mehta2017-vnect}   & 79.4 & 41.6 & - \\
    TrajectoryPose3D \cite{lin2019-trajectory}  & 83.6 & 51.4 & 79.8 \\
    UGCN \cite{wang2020-motion} & 86.9  & 62.1 & 68.1 \\
    U-CondDGCN \cite{hu(2021)-conditional} & 97.9   & 69.5 & 42.5 \\
    MixSTE~\cite{ZhangCVPR22MixSTE}(T=27)  & 94.4   & 66.5 & 54.9 \\ \hline
    HDFormer(T=32)           & 96.8  & 64.0   & 51.5 \\
    HDFormer(T=96)           & \textbf{98.7}  & \textbf{72.9}   & \textbf{37.2} \\ \hline
    \end{tabular}}
    \label{table:MPI}
    \vspace{-4mm}
\end{table}

\subsection{Qualitative Results}
\begin{figure}[!t]
    \centering
    \includegraphics[width=\linewidth]{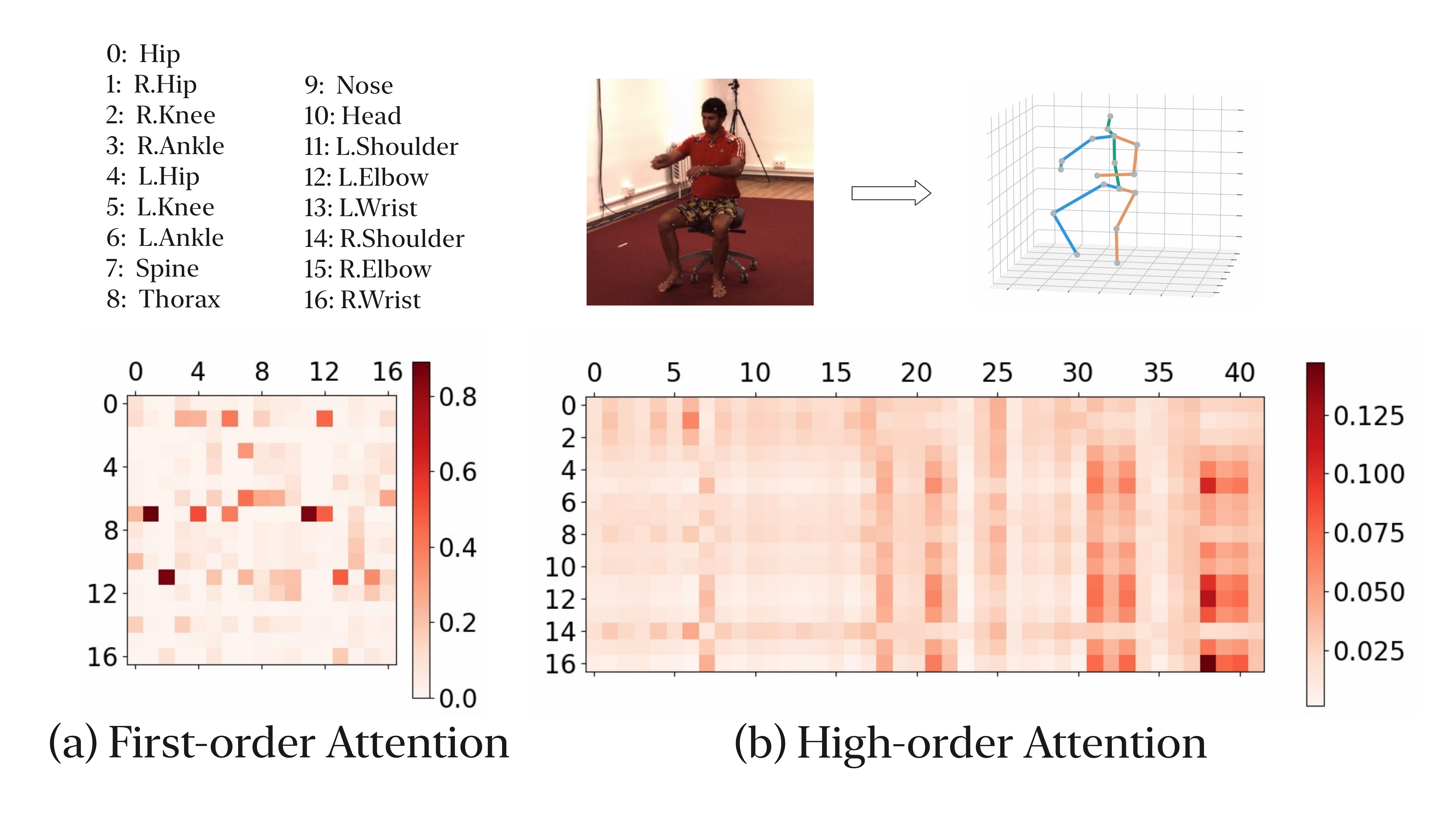}
    \vspace{-6mm}
    \caption{\small Visualization of first-order attentions and high-order attention between body joints and hyperbone.}
    \vspace{-1mm}
    \label{fig:vis_attn}
\end{figure}

As shown in Figure \ref{fig:vis_attn}, we further conduct visualization on the First-order attention and High-order attention. The selected action (Eating of test set S9) is applied for visualization. For the First-order attention map in Figure \ref{fig:vis_attn}(a), the horizontal and vertical axes are all joint indexes, and it can be easily observed that the dependency between the \textbf{spine} node and left/right \textbf{elbow} nodes are significant for the “eating” sequence. Besides, the left \textbf{shoulder} node also plays an important role in the spatial relationship with the left \textbf{ankle} node when eating in the sitting pose. So the first-order attention with self-attention of joints can capture the joint spatial relationship effectively.

Furthermore, to demonstrate the effect of the proposed high-order attention block, we further visualize the high-order attention map for the action of eating from the test set S9 in Figure \ref{fig:vis_attn}(b), where the vertical axes were the index of the joint while the horizontal axes were the index of hyperbones. In our experiments, the maximum SPD length was 4. As a result, the hyperbone sequence has 42 bone features in the horizontal axes. From the attention map, we can find that the hyperbone feature has an impact on different joints. The left/right \textbf{elbows} and left/right \textbf{wrist} have a relatively large response to hyperbone sequence index from 38 to 41, which corresponds to the higher order bone feature. The hyperbone 38 to 41 corresponding to joint sets (0,7,8,9,10), (0,7,8,11,12), (0,7,8,14,15), (7,8,11,12,13), which maps to the upper body parts and head. Besides, the \textbf{knee} joints are crooked when eating in a sitting pose as Figure \ref{fig:vis_attn} (b).

\begin{figure}[!t]
    \centering
    \includegraphics[width=\linewidth]{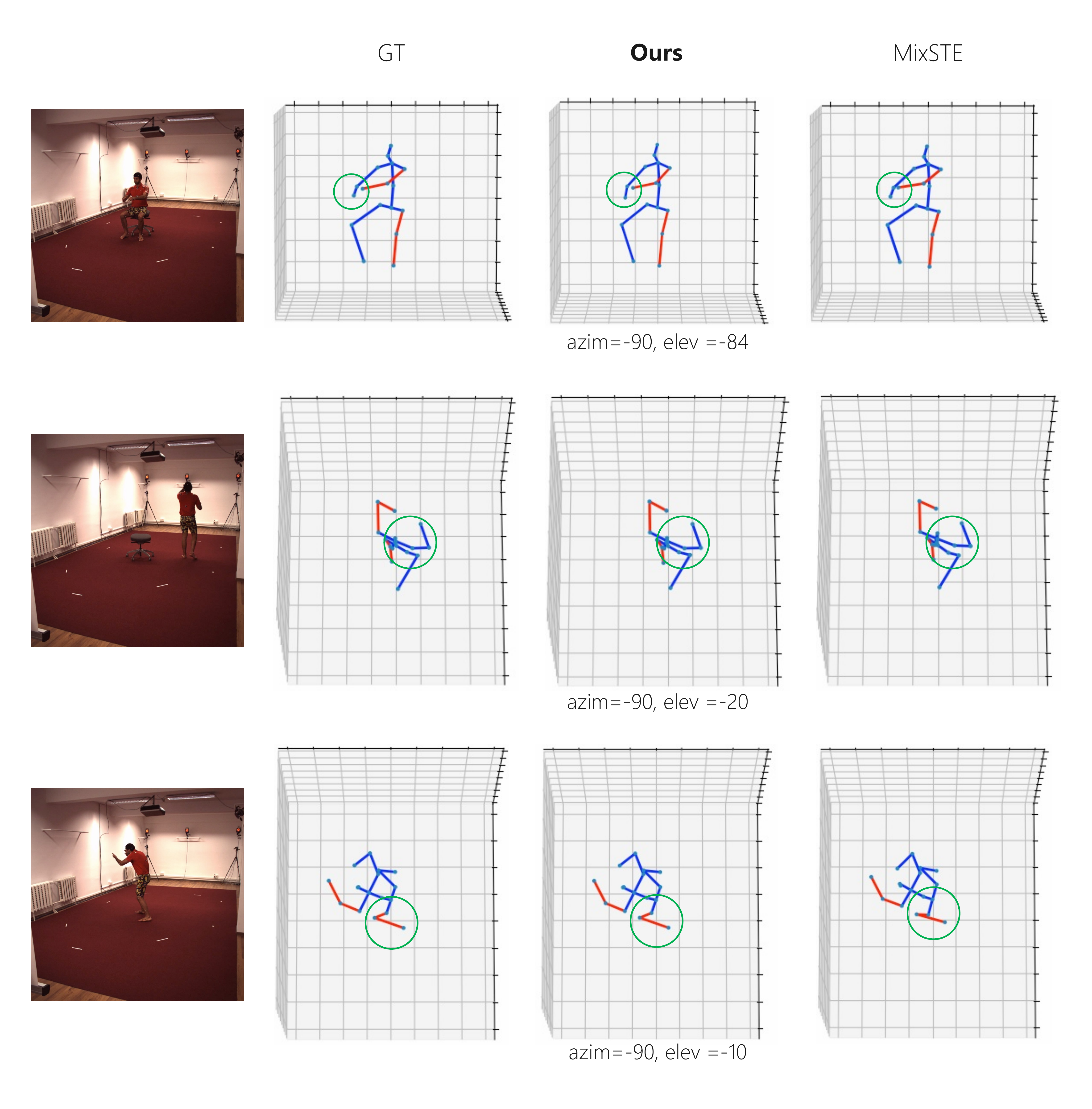}
    \vspace{-10mm}
    \caption{\small Qualitative comparison between our method (HDFormer) and \protect\cite{ZhangCVPR22MixSTE} with the Eating (first and second row) and Photo (third row) actions on Human3.6M. The green circle highlights locations where our method has better results.}
    \vspace{2mm}
    \label{fig:vis_compare}
\end{figure}

Besides, we also evaluate the visual result of estimated poses in Figure \ref{fig:vis_compare}. It can be seen that HDFormer estimates more accurate poses in cluttered and self-occlusion hands and feet compared to MixSTE \cite{ZhangCVPR22MixSTE}.

\subsection{Ablation Study}

\noindent \textbf{Impact of Multi-Order Attention.}~We evaluated the influence of our multi-order attention schema by conducting ablation studies on various order combinations, as shown in Table~\ref{table:ablation_order_number}. The experiments, with Human3.6M 2D ground truth key points as input, indicated an improvement in the Mean Per Joint Position Error (MPJPE) as the order number increased, with the best performance observed at $order=4$. This affirms the value of high-order attention in capturing complex skeletal information through "hyperbone$\leftrightarrow$joint" feature interaction.

\begin{table}
    \centering
    \scriptsize
    \caption{\small Ablation study of the order number. We compared the results of different orders involved in the high-order attention transformer block in Human3.6M with ground truth as input.}
    \vspace{-0.1in}
    \setlength{\tabcolsep}{1.55mm}{
    \begin{tabular}{l|c|c}
    \hline
    Methods         & Order & MPJPE[↓] \\ \hline
    \multirow{5}{*}{HDFormer}        & 1     & 25.0 \\
            & 2     & 23.6 \\
            & 3     & 22.8 \\
            & 4     & \textbf{21.6} \\
            & 5     & 22.7 \\
    \hline
    \end{tabular}}
    \label{table:ablation_order_number}
    \vspace{-4mm}
\end{table}

\begin{table}
    \centering
    \scriptsize
    \caption{\small Ablation study of the effectiveness of HDFormer at different stages. We compared the results of the baseline (UGCN~\protect\cite{wang2020-motion}), our HDFormer, and different configurations for our HDFormer on Human3.6M. The $\Delta$ denotes the improvements compared with the baseline.}
    \setlength{\tabcolsep}{1.55mm}{
    \begin{tabular}{l|c c c}
    \hline
    Methods        & High-order Attention & MPJPE[↓] & $\Delta$ \\ \hline
    Baseline & -  & 25.6 & - \\
    \hline
    \multirow{4}{*}{HDFormer} & Upsampling stage  & 24.1 & 1.5 \\
    & Downsampling stage  & 32.5 & -6.9 \\
    & Merging stage  & \textbf{21.6} & \textbf{4.0} \\
    & All stage  & 29.9 & -3.3 \\
    \hline
    \end{tabular}}
    \label{table:ablation_stages}
\end{table}

\noindent \textbf{Effectiveness of HDformer Block at Different Stages.} To explore the effectiveness of hyperbone feature and our proposed high-order transformer block, we conducted ablation studies on the Human3.6M dataset. To eliminate the influence of the 2D pose estimator, we adopted the ground-truth 2D poses as input. To explore the best stage to utilize the HDFormer block, we conducted experiments by adopting HDFormer block at various stages, and show the results in the bottom part of Table \ref{table:ablation_stages}. We found that adopting HDFormer block at the merging stage achieves a better result than other stages. The reason behind this could be the complex skeleton dynamics of hyperbones can not be learned at early stages (i.e., downsampling stage), therefore, leads to inferior performance. In the merging stage, HDFormer block can fuse complex information from previous stages and build high-order aggregation of skeleton structure information by high-order attention. Compared with adopting HDFormer at the downsampling stage (yields 6.9mm decrease), adopting it at all stages (yields 3.3mm decrease) and adopting it at the merge stage get the best performance (yields 4.0 mm improvement).

\begin{table}
    \centering
    \scriptsize
    \caption{\small Ablation study of hyperbone representation. We compared the results of different edge feature representations for hyperbone encoding in Human3.6M with ground truth as input.}
    \vspace{-0.1in}
    \setlength{\tabcolsep}{1.55mm}{
    \begin{tabular}{l|c|c|c}
    \hline
    Methods        & hyperbone representation & MPJPE[↓] & $\Delta$ \\ \hline
    Baseline     & -                & 25.6 & - \\
    \hline
    \multirow{4}{*}{HDFormer} & summation        & 23.5 & 2.1 \\
     & multiplication   & 23.1  & 2.5 \\
     & concatenation    & 23.5 & 2.1 \\  
     & subtraction + concatenation      & \textbf{21.6} & 4.0 \\
    \hline
    \end{tabular}}
    \label{table:ablation_feature_representation}
    \vspace{-2mm}
\end{table}

\begin{table}
    \tiny
    \centering
    \caption{\small More results on Human3.6M with GT 2D poses as input.}
    \vspace{-3mm}
    \setlength{\tabcolsep}{1.35mm}{
    \begin{tabular}{l|c|c|c|c|c}
    \hline
    Line & Methods                         & MPJPE[↓]  & Params & Frames \\ \hline
    1 & HDFormer (w/o $\Psi$)          & 27.4 & 3.7 M &  96\\
    2 & HDFormer (with pos encoding)    & 22.1 & 3.8 M &  96\\
    3 & HDFormer (multi-head concat)    & 22.9 & 3.7 M &  96\\
    4 & HDFormer (T=243)                & 21.8 & 4.7 M &  96\\
\hline
    5 & HDFormer (proposed)             & \textbf{21.6} & 3.7 M &  96 \\ \hline
    \end{tabular}}
    \label{table:H36M_GT2d_ablation}
    \vspace{-2mm}
\end{table}

\noindent \textbf{Exploration of Hyperbone Representation.} The hyperbone representation is a vital factor for the graph skeleton structure, and we exploit the way of the hyperbone feature representation. We conducted experiments by adopting $4^{th}$ order HDFormer block with different instantiations modes, which includes summation, multiplication, concatenation, and subtraction+concatenation as can be shown in Table~\ref{table:ablation_feature_representation}. Compared to the baseline method, we found that all the hyperbone representation methods outperform the baseline, as they utilize high-order information. Among them, subtraction + concatenation boosts the performance over baseline by 4.0mm. It shows that bone feature concatenation with shortest path aggregation is effective for hyperbone feature representation.

\noindent \textbf{Role of Position Encoding and Multi-Head Attention.} We observed from our experimental results (line 2 of Table \ref{table:H36M_GT2d_ablation}) that incorporating absolute positional encoding led to a decrease in performance. This suggests that position coding is not helpful in improving performance. Besides, we have conducted an ablation study on the use of concatenation and summation in the multi-head attention module, and we found that summation resulted in better performance, which can be shown in line 3 of Table \ref{table:H36M_GT2d_ablation}. Consequently, we adopted the summation in the multi-head attention in our proposed model.

\noindent \textbf{Longer Frames as Input.} Extending the input frame numbers to 243 led to a slight decline compared to 96 frames when using 2D ground truth input as shown in line 4 of \ref{table:H36M_GT2d_ablation}. We suspect that the reason might be the small scale of our model (1/10 compared to \cite{ZhangCVPR22MixSTE}) may not well capture temporal redundancy and noise in dense sequence.

\section{Conclusion}
In this work, we propose a novel model named High-order Directed Transformer (HDFormer), which considers both ``\textit{joint$\leftrightarrow$joint}", ``\textit{bone$\leftrightarrow$joint}" and ``\textit{hyperbone$\leftrightarrow$joint}" connections. Specifically, we propose a hyperbone representation learning module and a high-order attention module to model complicated semantic relations between hyperbone and joint. We conduct extensive experiments to provide both quantitative and qualitative analysis. Our proposed method achieves state-of-the-art performances with only 1/10 parameters and a fraction of computational cost compared to recently published SOTA.

\section*{Acknowledgments}
{\small The research work of Zhi-Qi Cheng in this project received support from the US Department of Transportation, Office of the Assistant Secretary for Research and Technology, under the University Transportation Center Program with Federal Grant Number 69A3551747111. Additional support came from the Intel and IBM Fellowships. The views and conclusions contained herein represent those of the authors and not necessarily the official policies or endorsements of the supporting agencies or the U.S. Government.}
\bibliographystyle{named}
\bibliography{ijcai23}

\end{document}